\documentclass{article}

\usepackage{microtype}
\usepackage{graphicx}
\usepackage{booktabs} 

\usepackage{hyperref}

\usepackage[accepted]{icml2024}

\usepackage{latexsym}
\usepackage{mathptmx} 

\usepackage{hyperref}
\usepackage{xcolor}

\usepackage{graphicx,epstopdf}			

\graphicspath{{graphics/}} 

\usepackage{natbib}

\usepackage{times}
\usepackage{latexsym}
\usepackage{url}

\usepackage{lipsum} 
\usepackage{stmaryrd}
\usepackage{bbm}

\usepackage{amsmath}
\usepackage{amsthm}
\usepackage{amsfonts}
\usepackage{bm}
\usepackage{thmtools,mathtools}
\usepackage{thm-restate}
\usepackage[T1]{fontenc}
\usepackage{booktabs}
\usepackage{multirow}
\usepackage{booktabs}
\usepackage{verbatim}
\usepackage{mdwlist}
\usepackage[ruled,norelsize,linesnumbered]{algorithm2e}
\usepackage{dsfont}
\usepackage[shortlabels]{enumitem}
\usepackage{caption}
\usepackage{subcaption}
\usepackage{array}
\usepackage{pifont}
\usepackage{CJKutf8}

\usepackage{multicol}
\usepackage{float}
\usepackage{lipsum}
\usepackage[font=small]{caption}
\usepackage{svg}
\usepackage{wrapfig}
\theoremstyle{definition}

\makeatletter
\newcommand{\removelatexerror}{\let\@latex@error\@gobble}
\makeatother

\newcommand{\codename}{\textit{Reprompting}\xspace}

\usepackage[textsize=tiny]{todonotes}

\icmltitlerunning{Reprompting: Automated Chain-of-Thought Prompt Inference Through Gibbs Sampling}

\begin{document}

\twocolumn[
\icmltitle{Reprompting: Automated Chain-of-Thought Prompt Inference\\Through Gibbs Sampling}




\begin{icmlauthorlist}
\icmlauthor{Weijia Xu}{comp}
\icmlauthor{Andrzej Banburski-Fahey}{comp}
\icmlauthor{Nebojsa Jojic}{comp}
\end{icmlauthorlist}

\icmlaffiliation{comp}{Microsoft Research, Redmond, USA}

\icmlcorrespondingauthor{Weijia Xu}{weijiaxu@microsoft.com}



\vskip 0.3in
]



\printAffiliationsAndNotice{}  

\begin{abstract}
  We introduce Reprompting, an iterative sampling algorithm that automatically learns the Chain-of-Thought (CoT) recipes for a given task without human intervention. Through Gibbs sampling, Reprompting infers the CoT recipes that work consistently well for a set of training samples by iteratively sampling new recipes using previously sampled recipes as parent prompts to solve other training problems. We conduct extensive experiments on~20 challenging reasoning tasks. Results show that Reprompting outperforms human-written CoT prompts substantially by~+9.4 points on average. It also achieves consistently better performance than the state-of-the-art prompt optimization and decoding algorithms.
\end{abstract}

\section{Introduction}
\looseness=-1
Few-shot prompting with large language models~(LLMs) has revolutionized the landscape of natural language processing. Given natural language instructions and a few demonstrations as in-context examples, LLMs can quickly adapt to new tasks, approaching or even surpassing the performance of models fine-tuned on larger datasets on a wide range of tasks~\citep{gpt3}. However, such prompting techniques fall short on tasks that require multi-step reasoning and constraint propagation~\citep{wei2022chain}, such as \textit{logical deduction} in the Big-Bench Hard benchmark~\citep{suzgun2022challenging}. To address these limitations, prior works proposed to teach LLMs to reason step by step like humans by prompting them with chain-of-thought~(CoT) reasoning steps for a few example problems~\citep{wei2022chain}. Despite the improved performance, such a method requires human experts with not only the task knowledge but also an understanding of how prompting works to craft the CoT prompt for each task~\citep{zamfirescupereira2023why}, which limits the scalability and generalizability of the method. Furthermore, a problem can be reasoned in many different ways, and some of them may work well on some LLMs but not on others. To fairly compare the performance of various LLMs on each task, we need to find the CoT prompt that works best for each model in a feasible way, which remains a challenge.

In this paper, we propose \codename, an iterative sampling algorithm that \textbf{automatically} finds effective CoT prompt for each model given a few question-answer pairs without human intervention. Specifically, the algorithm aims to infer a set of CoT recipes that perform consistently well as in-context examples for a set of training problems. We frame it as a problem of sampling from a joint distribution of CoT recipes given the training question-answer pairs, which is infeasible to characterize directly but can be approached using Gibbs sampling \---\ we initially sample a set of recipes through zero-shot prompting, expand the set with new recipes sampled iteratively by using previously sampled recipes as parent prompts to solve a different training problem, and weed out the least-fit recipes that lead to wrong answers. Thus, the algorithm will eventually converge to a set of recipes that share similar chains of thought for effectively solving the training problems. These CoT recipes optimized on the training set then serve as effective CoT prompts for solving unseen test problems.

We evaluate \codename on~20 tasks from three reasoning benchmarks including Big-Bench Hard~(BBH)~\citep{suzgun2022challenging}, GSM8K~\citep{cobbe2021gsm8k} and MATH~\citep{hendrycks2021math} using ChatGPT~\citep{openai2023gpt4} and InstructGPT~\citep{ouyang2022training} as LLMs.
Compared with human-written CoT prompts, \codename achieves~+9.4 higher accuracy on average. It also consistently outperforms self-consistency decoding~\citep{wang2022self}, Auto-CoT~\citep{zhang2022automatic} and Automatic Prompt Optimization~\citep{pryzant2023automatic} by~11--33 points on average. Furthermore, \codename facilitates model combination by using different LLMs for initializing and sampling new recipes. Empirically, leveraging ChatGPT to sample initial recipes for InstructGPT brings up to~+71 point improvements over using InstructGPT alone and even outperforms ChatGPT alone on certain tasks. Lastly, our results confirm that the CoT recipes that work well on one model may work poorly on another, even when the latter may approach the best performance using prompts optimized for itself. These findings emphasize the need to optimize the prompt for each model for fair comparisons.

\section{\codename: Prompt Inference Through Gibbs Sampling}
\subsection{In-Context Learning}
 \textbf{In-context learning} has become the cornerstone of evaluating large language models (LLMs) \citep{gpt3,srivastava2022beyond}. To facilitate this evaluation approach, data is provided for a large number of different tasks, with each task consisting of dozens or, more often, hundreds of instances with varying problem setup and question texts $x_i$ and their corresponding text answers $y_i$, where $i\in [1..N]$ and $N$ is the number of problem instances for the task. Formally, in-context learning infers the answer for a given test question~$x$ by prompting an LLM with a set of demonstration examples~$\{x_i,y_i\}_{i=1}^K$:
\begin{equation}
    \hat{y} \sim p_{LLM}(y | \{x_i,y_i\}_{i=1}^K, x)
\end{equation}

The performance of in-context learning can be significantly enhanced by incorporating auxiliary knowledge or human-written instructions in a prompt~\citep{shwartz-etal-2020-unsupervised,zelikman2022star,nye2021show}, particularly in the form of Chain-of-Thought (CoT) reasoning~\citep{wei2022chain, wang2022self, zhou2022least, creswell2022selection, wang2022rationale,liu-etal-2022-multi, kojima2022large, li2022advance}.

\begin{figure}[t]
    \centering
    \includegraphics[width=.48\textwidth]{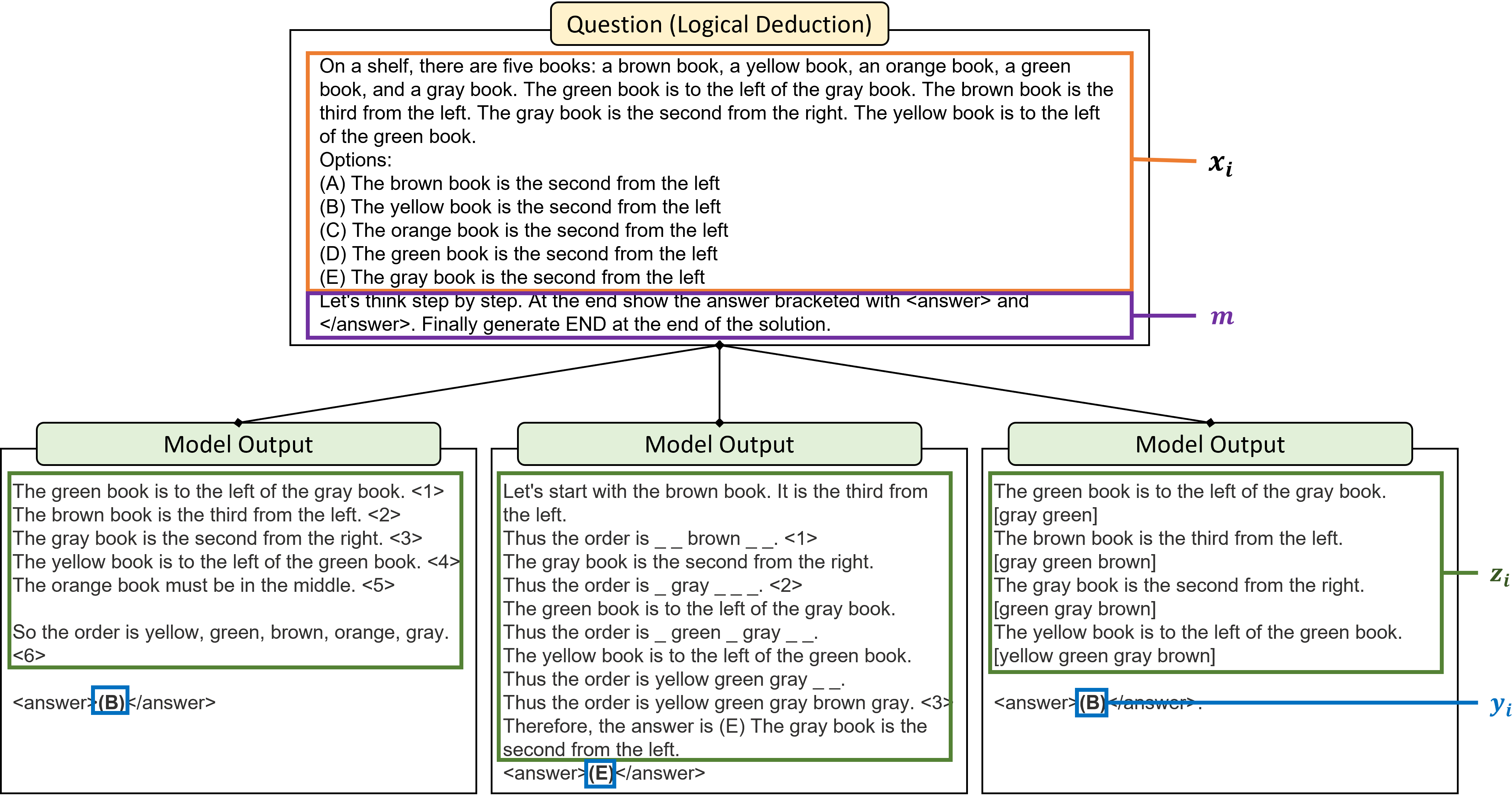}
\caption{An example that ChatGPT can propose various different solutions to the same problem in zero-shot.}
\label{fig:zeroshot_solutions}
\end{figure}

In-context learning with CoT~\citep{wei2022chain} can be seen in a similar light, statistically. In addition to the question-answer pairs~$\{x_i, y_i\}$, the CoT prompt also contains worked out step-by-step reasoning ``recipes''~$z_i$ in text, which are inserted between the question and answer:~$\{x_i, z_i, y_i\}$. These recipes can play two roles. First, they further explain the intent of the question~$x_i$, as a small collection of question-answer pairs alone may be insufficient to disambiguate among different patterns an LLM might detect. The second role is more important: it provides step-by-step guidance on one problem and thus teaches an LLM to solve similar problems following the same routine as it continues the text conditioned on the previous tokens.
In the extreme, with prompts that strictly regiment self-attention, GPT models can be turned into Turing Machines to execute standard computer algorithms \citep{jojic2023gptTM}. In practice, the CoT prompts commonly used in prior work fall somewhere between colloquial explanations and regimented recipes.
Formally, {\bf in-context learning with CoT} infers the answer for a given test question~$x$ by prompting an LLM with an optional instruction message~$m$ and a set of demonstration examples with step-by-step solutions~$\{x_i, z_i, y_i\}_{i=1}^K$:
\begin{equation}
    \hat{z}, \hat{y} \sim p_{LLM}(z, y | \{x_i, z_i, y_i\}_{i=1}^K, x, m)
\label{eq:in_context_assumption}
\end{equation}

Here, $m$ is a textual message that instructs the model to generate the step-by-step solution~$z_j$ before the answer text~$y_j$ and the specific format to present the answer.\footnote{This enables us to separate the generated answer~$y_j$ from the step-by-step solution~$z_j$ and forces the model to stop after generating the answer.} It can be task-specific or generic, as in the case of our experiments. Such an instruction message can trigger instruction-tuned LLMs to generate step-by-step solutions given~$[x_j, m]$ alone without any demonstration examples~(i.e.~$K = 0$), as illustrated in Figure~\ref{fig:zeroshot_solutions}. These solutions follow varying styles and often lead to incorrect answers. However, we argue that good recipes for solving the set of problems on a given task can evolve from these zero-shot solutions. In the next section, we introduce \codename, an iterative sampling algorithm that automatically produces the CoT recipes for a given set of problems without human intervention.

\subsection{Prompt Inference Through Gibbs Sampling}
We introduce the \codename algorithm, which aims to find a set of CoT recipes~$z_i$ that work {\bf consistently} well as few-shot in-context examples for a dataset $\{x_i,y_i\}_{i=1}^N$. Specifically, we formulate it as the problem of sampling from a joint distribution
\begin{equation}
    p(z_1, z_2,...z_N| \{x_i,y_i\}_{i=1}^N, m)
\label{eq:joint_distribution}
\end{equation}
such that~$z_{1...N}$ are generalized enough so that given any test question~$x$, the distribution over~$z$ and~$y$ is approximately invariant to the choice of the $K$-shot CoT recipes:
\begin{equation}
\begin{split}
    & p_{LLM}(z, y | \{x_i, z_i, y_i\}_{i=1}^N, x, m) \\
    \approx& p_{LLM}(z, y | \{x_i, z_i, y_i\}_{i \in S}, x, m), \, \quad \forall S \subset [1,N], |S| = K
\end{split}
\label{eq:reprompting_approximation}
\end{equation}

Without characterizing the joint distribution, we can use Gibbs sampling~\citep{Geman1984StochasticRG} to generate such samples~$\{z_1, z_2,...z_N\}$ by first sampling~$\{z_1, z_2,...z_N\}$ independently from the distributions~$p(z_j | x_j, y_j)$, and then iteratively drawing samples from the conditional distributions~$p(z_j | z_1,...,z_{j-1},z_{j+1},...z_N, \{x_i,y_i\}_{i=1}^N, m)$. Based on the property~\eqref{eq:reprompting_approximation} of the joint distribution, we have the following approximation:
\begin{equation}
\begin{split}
    &p(z_j | z_1,...,z_{j-1},z_{j+1},...z_N, \{x_i,y_i\}_{i=1}^N, m) \\
    =& p_{LLM}(z_j | \{x_i, z_i, y_i\}_{i \neq j}, x_j, y_j, m) \\
    \propto& p_{LLM}(z_j, y_j | \{x_i, z_i, y_i\}_{i \neq j}, x_j, m) \\
    \approx& p_{LLM}(z_j, y_j | \{x_i, z_i, y_i\}_{i \in S_j}, x_j, m), \, \\
    &\forall S_j \subset [1,N]\backslash \{j\}, |S_j| = K
\end{split}
\end{equation}
Thus, we can sample~$z_j$ by randomly picking $K$ data points~(excluding~$j$) and then sampling~$z_j$ with weights proportional to the conditional probability
\begin{equation}
\begin{split}
    &p_{LLM}(z_j, y_j | \{x_i, z_i, y_i\}_{i \in S_j}, x_j, m) \\
    =& p_{LLM}(z_j | \{x_i, z_i, y_i\}_{i \in S_j}, x_j, m) \\
    &\cdot p_{LLM}(y_j | \{x_i, z_i, y_i\}_{i \in S_j}, x_j, m, z_j)
\end{split}
\end{equation}
One way to approximate it is to sample several~$\hat{z}_j$ from the LLM conditioned on~$\{x_i, z_i, y_i\}_{i \in S_j}$, $x_j$ and~$m$, compute the weight for each~$\hat{z}_j$ using the model's probability of the correct answer~$y_j$ conditioned on~$\{x_i, z_i, y_i\}_{i \in S_j}$, $x_j$,~$m$ and~$\hat{z}_j$, and sample a~$z_j$ from~$\{\hat{z}_j\}$ based on the weights. In practice, however, the model likelihood of a given text may be inaccessible. Thus, we approximate it using rejection sampling \---\ we sample~$z_j$ by sampling~$\hat{z}_j$ and~$\hat{y}_j$ from~$p_{LLM}(z, y | \{x_i, z_i, y_i\}_{i \in S_j}, x_j, m)$ and then reject~$\hat{z}_j$ with a probability of~$p_{rej}$ if~$\hat{y}_j \neq y_j$. Otherwise, we accept~$\hat{z}_j$ and update the sample. 
Algorithm~\ref{alg:reprompting} shows the complete \codename algorithm consisting of the initialization and iterative sampling steps. Note that we set the rejection probability~$p_{rej}$ in a way that allows solutions that lead to incorrect answers to be kept occasionally, as these solutions may still contain useful segments that evolve into good recipes through \codename.

\begin{algorithm}[ht]
\DontPrintSemicolon
\SetKwInOut{Input}{Input}
\Input{Training set $\{x_i,y_i\}_{i=1}^N$, number of examples in the prompt $K$, number of iterations $M$, rejection probability~$p_{rej}$, the initialization model~$LLM_1$ and the sampling model~$LLM_2$}
\textbf{Initialization:}\;
\For {each $j$}{
$z_j \leftarrow \emptyset$\;
Sample $\hat{z}_j, \hat{y}_j \sim p_{LLM_1}(z, y | x_j, m)$\;
Sample $u \sim Uniform([0, 1])$\;
\If {$\hat{y}_j = y_j$ or $u>p_{rej}$}{
$z_j \leftarrow \hat{z}_j$\;
}
}
\textbf{Sampling:}\;
\Repeat{convergence or $M$ iterations are reached}{
Randomly select $j \in [1, N]$\;
Randomly select $S_j \subset [1,N]\backslash \{j\}$ of size $K$\;
Sample $\hat{z}_j, \hat{y}_j \sim p_{LLM_2}(z, y | \{x_i, z_i, y_i\}_{i \in S_j}, x_j, m)$\;
Sample $u \sim Uniform([0, 1])$\;
\If {$\hat{y}_j = y_j$ or $u>p_{rej}$}{
$z_j \leftarrow \hat{z}_j$\;
}
}
\caption{\codename algorithm}
\label{alg:reprompting}
\end{algorithm}

Based on the properties of Gibbs sampling~\citep{casella1992explaining,roberts1994simple}, the algorithm should converge to the point where the probability $p_{LLM}(z_j, y_j | \{x_i, z_i, y_i\}_{i \in S_j}, x_j, m)$ is high and agnostic to the choice of~$S_j$, which leads to a set of~$\{z_j\}$ that work well as a prompt for solving similar problems in a separate test set.

The algorithm can also be viewed as a variant of evolutionary algorithms:~1) First, we generate the initial population of individuals (where each individual is a CoT recipe given a problem). 2) Next, we repeat the following regeneration steps iteratively: 2a) we first evaluate the fitness of each CoT recipe by comparing the answer that follows the recipe with the correct answer and weed out the least-fit recipes; 2b) we then breed new individuals through crossover and mutation by randomly selecting K recipes from the population as parent recipes, which are then used to prompt the LLM to generate recipes for a new problem. By repeating the 2a and 2b steps, initial recipes can be recombined~(Figure~\ref{fig:reprompt_combine_examples}) and evolve into better recipes (Figure~\ref{fig:reprompt_examples}) through iterations. And eventually, the fittest recipes (i.e. ones that can be followed to solve similar problems) will survive.

During testing, we select~$K$ tuples~$\{x_i, z_i, y_i\}$ from the inferred~$\{z_j\}$ based on the training accuracy when using each tuple individually in a prompt.

\section{Experimental Setup}
We evaluate the \codename algorithm against various baselines including zero-shot, few-shot, Chain-of-Thought~(CoT), Chain-of-Thought combined with self-consistency decoding~\citep{wang2022self}, Auto-CoT~\citep{zhang2022automatic} and Automatic Prompt Optimization~\citep{pryzant2023automatic} on~20 challenging reasoning tasks, including~12 challenging tasks in the Big-Bench Hard~(BBH) benchmark~\citep{suzgun2022challenging},\footnote{The BBH tasks include \textit{Logical Deduction}, \textit{Geometric Shapes}, \textit{Object Counting}, \textit{Penguins in a Table}, \textit{Temporal Sequences}, \textit{Date Understanding}, \textit{Formal Fallacies}, \textit{Movie Recommendation}, \textit{Reasoning About Colored Objects}, \textit{Ruin Names}, \textit{Salient Translation Error Detection}, and \textit{Word Sorting}.} GSM8K~\citep{cobbe2021gsm8k} and MATH~\citep{hendrycks2021math}.
We choose both tasks that have been shown to benefit substantially from human-written CoT recipes, such as Logical Deduction, Geometric Shapes, Temporal Sequences, GSM8K and MATH, and tasks on which CoT does not improve much or does not improve consistently over zero-shot prompting, such as Formal Fallacies, Movie Recommendation and Word Sorting.

\subsection{\codename Setup} 
\looseness=-1
For each task, we randomly select 20 training examples from the Big-Bench dataset excluding the test examples in the BBH benchmark.\footnote{Except for \textit{Penguins in a Table} where there are only three samples in the Big-Bench dataset that are excluded from BBH, so we randomly select~17 more examples from BBH into the training set.} We experiment with having~$k \in \{1, 3\}$ clones of the same training example in the set~$\{x_i,y_i\}_{i=1}^N$ to allow for more diverse recipe samples~(so the number of recipes we need to sample from the joint distribution~\eqref{eq:joint_distribution} is~$N = 20*k$) and choose~$k$ that obtains the highest training accuracy. 
We set the number of examples in the prompt by~$K = 5$. We run \codename for a maximum of~$M = 20,000$ iterations. We allow for early stopping if the average training accuracy stops increasing for~$1,000$ iterations. For the rejection probability, we experiment with~$p_{rej} \in \{0.95, 0.99\}$ and choose~$p_{rej} = 0.99$ as it leads to higher training accuracy on various tasks. 

\subsection{Baselines} 
\paragraph{Prompting Baselines}
For \textbf{zero-shot prompting}, we only include the test question~$x_i$ and the special message~$m$ in the prompt, which triggers the model to generate a step-by-step solution prior to the answer text. For \textbf{few-shot prompting}, we randomly select 20 training examples in the same way as in \codename and concatenate these examples in the form of question-answer pairs in the prompt, followed by the test question. For \textbf{CoT prompting}, we use the human-written CoT prompts from \citet{suzgun2022challenging}. For \textbf{CoT with self-consistency decoding}, we use the same CoT prompts and follow \citet{wang2022self} by sampling~10 reasoning paths per question and taking the majority vote on the answer. For both approaches, we randomly select~20 training examples in the same way as in \codename.\footnote{Recent prompting methods that are more annotation-intensive, such as Complex-CoT~\citep{fu2022complexity} and Progressive-Hint Prompting~\citep{zheng2023progressive}, are shown to outperform \codename by 3.3--5.6 points on GSM8K. However, these methods leverage substantially more human-annotated examples~(e.g. 7.5K annotated examples on GSM8K) than \codename, thus they are not directly comparable.}

\paragraph{Prompt Optimization Baselines}
We also compare \codename with two previous state-of-the-art prompt optimization algorithms, including \textbf{Auto-CoT}~\citep{zhang2022automatic} and \textbf{APO}~\citep{pryzant2023automatic}.
For \textbf{Auto-CoT}, since the original Auto-CoT algorithm differs from our setting as it focuses on the unsupervised setting without exploiting any labeled examples, we adapt the algorithm to our few-shot setting where it follows the original algorithm to generate diverse CoT recipes through zero-shot prompting but selects the demonstration examples based on the training accuracy when used individually in a prompt.\footnote{The original Auto-CoT algorithm selects the demonstration examples based on the diversity of the demonstration questions.} We also evaluate \textbf{APO}, a recently proposed nonparametric prompt optimization algorithm that uses LLMs to generate ``textual gradient'' \---\ criticism of the current prompt \---\ based on training samples and edit the prompt accordingly. The algorithm has been shown to outperform other prompt optimization methods, such as TEMPERA~\citep{zhang2023tempera}, Automatic Prompt Engineering~\citep{zhou2023large}, and AutoGPT.\footnote{\url{https://news.agpt.co/}}

\subsection{Large Language Models~(LLMs)} 
We experiment with two powerful LLMs including ChatGPT~(gpt-3.5-turbo; \citet{openai2023gpt4}) and InstructGPT~(text-davinci-003; \citet{ouyang2022training}). We also experiment with a combo model for \codename where we use ChatGPT as~$LLM_1$ for initialization and InstructGPT as~$LLM_2$ for sampling. For both LLMs, we set the maximum number of output tokens to~500, $top\_p = 0.5$, zero frequency and presence penalty. Additionally, we include ``END'' as the stop word. We set the temperature to~1.0 for \codename and~$0.0$ for testing.

\subsection{Evaluation Protocol} 
We extract the final answer from the model output by extracting the text between ``<answer>'' and ``</answer>'', except for the CoT baseline where we extract the final answer in the same way as in \citet{suzgun2022challenging}. We measure accuracy based on exact match by comparing the extracted answer with the ground truth.
\begin{table*}[ht]
\centering
    \scalebox{0.9}{
    \begin{tabular}{l|c|cccccc|ccc}
\multirow{2}{*}{ BBH Task } & \multirow{2}{*}{ SOTA } & ZS & FS & CoT & CoT+SC & APO & AutoCoT & \multicolumn{3}{c}{\codename} \\
& & \multicolumn{6}{c|}{ChatGPT} & ChatGPT & InsGPT & Chat+Ins \\
\midrule
Logical & 60.4 & 35.1 & 46.4 & 63.1 & 62.7 & 28.0 & 53.2 & \textbf{66.3} & 53.7 & 60.0 \\
Geometric & 56.0 & 13.6 & 20.0 & 58.0 & 60.0 & 52.0 & 52.4 & \textbf{72.8} & 40.8 & 64.4 \\
ObjectCount & 93.2 & 52.4 & 46.8 & 95.6 & 95.2 & 74.8 & 88.8 & 97.2 & 42.8 & \textbf{99.6} \\
Penguins & 81.5 & 50.7 & 60.3 & 67.1 & 71.2 & 45.2 & \textbf{85.6} & \textbf{85.6} & 78.1 & 82.9 \\
Temporal & 96.8 & 38.4 & 41.2 & 66.8 & 66.8 & 50.4 & 80.8 & 93.2 & 28.4 & \textbf{99.2} \\
\midrule
\textbf{Average} & 77.6 & 38.0 & 42.9 & 70.1 & 71.2 & 50.1 & 72.2 & \textbf{83.0} & 48.8 & 81.2 \\
\midrule
\end{tabular}}
\caption{Performance of several large language models~(LLMs) using \codename versus the baseline prompting and prompt optimization methods on Big-Bench Hard~(BBH) tasks. \textit{SOTA} refers to the state-of-the-art performance among InstructGPT~(text-davinci-002; \citet{ouyang2022training}), Codex~\citep{copilot}, and PaLM 540B~\citep{chowdhery2022palm} using CoT prompting from \citet{suzgun2022challenging}. We also compare \codename with ChatGPT using \textit{ZS}~(zero-shot), \textit{FS}~(few-shot), \textit{CoT}, \textit{CoT+SC} (CoT prompting combined with self-consistency decoding~\citep{wang2022self}), \textit{APO} (automatic prompt optimization using textual gradient~\citep{pryzant2023automatic}), and \textit{AutoCoT} (the few-shot version of Auto-CoT~\citep{zhang2022automatic}). For \codename, we show the performance of various LLMs \---\ including \textit{ChatGPT}~(gpt-3.5-turbo; \citet{openai2023gpt4}), \textit{InstructGPT}~(text-davinci-003), and \textit{Chat+Instruct} (a combo version that uses ChatGPT for initialization and InstructGPT at sampling steps).}
\label{tab:main_results}
\end{table*}

\begin{table}[ht]
\centering
\begin{tabular}{l|ccc|c}
& ZS & FS & CoT & \codename \\
\midrule
\textbf{BBH} & & & & \\
Date & 63.6 & 46.4 & \textbf{76.8} & 76.4 \\
Formal & 49.2 & 53.6 & 48.4 & \textbf{56.8} \\
Movie & 59.2 & 72.4 & 25.6 & \textbf{78.4} \\
ColoredObj & 66.8 & 48.8 & \textbf{76.0} & 74.0 \\
Ruin & 53.2 & 66.8 & 60.8 & \textbf{74.8} \\
Salient & 43.2 & 53.2 & 32.8 & \textbf{54.8} \\
WordSort & 58.0 & 72.0 & 46.0 & \textbf{73.2} \\
\midrule
\textbf{GSM8K} & 45.6 & 26.5 & 75.6 & \textbf{79.5} \\
\midrule
\textbf{MATH} & & & & \\
Algebra & 37.6 & 23.7 & 52.0 & \textbf{53.1} \\
Counting & 17.1 & 19.8 & 26.6 & \textbf{32.3} \\
Geometry & 12.4 & 16.2 & 28.5 & \textbf{29.2} \\
IntAlgebra & 9.4 & 12.1 & \textbf{18.0} & 16.8 \\
Number & 20.8 & 17.1 & 32.9 & \textbf{33.3} \\
Prealgebra & 31.4 & 33.2 & \textbf{54.0} & 43.8 \\
Precalculus & 7.4 & 18.4 & 19.0 & \textbf{19.3} \\
\midrule
\textbf{Average} & 38.3 & 38.7 & 44.9 & \textbf{53.0} \\
\midrule
\end{tabular}
\caption{Performance of ChatGPT using \codename versus \textit{ZS}~(zero-shot), \textit{FS}~(few-shot), and \textit{CoT} prompting methods on seven additional tasks from Big-Bench Hard~(BBH)~\citep{suzgun2022challenging}, GSM8K~\citep{cobbe2021gsm8k} and MATH~\citep{hendrycks2021math}.}
\label{tab:additional_results}
\end{table}

\begin{table}[t]
\centering
\begin{tabular}{l|ccc|c}
& $p_{rej}=0$ & $p_{rej}=1$ & NoRec & Orig. \\
\midrule
Logical & 56.3 & 61.9 & 54.7 & \textbf{66.3} \\
ObjectCount & 52.0 & \textbf{97.2} & 95.6 & \textbf{97.2} \\
Temporal & 74.8 & 74.4 & 90.4 & \textbf{93.2} \\
\midrule
\textbf{Average} & 61.0 & 77.8 & 80.2 & \textbf{85.6} \\
\midrule
\end{tabular}
\caption{Ablation study on rejection sampling (including no rejection~($p_{rej}=0$) and always rejecting~($p_{rej}=1$)) and recombination~(\textit{NoRec} represents \codename without recombination of previously sampled recipes) on Logical Deduction, Object Counting, and Temporal Sequences from Big-Bench Hard~(BBH)~\citep{suzgun2022challenging}. The \textit{Orig.} column represents the standard \codename algorithm without ablation.}
\label{tab:ablation}
\end{table}

\begin{table}
    \centering
    \begin{tabular}{l|ll}
    Tasks & InsGPT & ChatGPT \\
    \midrule
    Logical & 65.9 & \textbf{66.3}$^*$ \\
    Geometric & 53.6 & \textbf{72.8}$^*$ \\
    ObjectCount & \textbf{99.6}$^*$ & 96.8 \\
    Penguins & 82.2 & \textbf{85.6}$^*$ \\
    Temporal & \textbf{99.2}$^*$ & 81.6 \\
    \midrule
\end{tabular}
\caption{Testing the best performing CoT prompt learned on ChatGPT, InstructGPT or InstructGPT+ChatGPT through \codename on both ChatGPT and InstructGPT. The superscript~$^*$ denotes the model used as~$LLM_2$ in \codename.}
\label{tab:test_across_models}
\end{table}

\section{Results}
\subsection{Main Results}
We first compare the performance of \codename with all the baselines on five BBH tasks. As shown in Table~\ref{tab:main_results}, results confirm the previous finding that few-shot in-context prompting improves the performance over zero-shot~\citep{gpt3} and that CoT prompting outperforms both zero-shot and few-shot prompting by a large margin. However, human-written CoT prompting requires costly prompt engineering, as not all CoT recipes work equally well on LLMs~\citep{madaan2022,jojic2023gptTM}. 
Crucially, we show that using \codename, LLMs can achieve better performance compared to the existing CoT prompts, but without requiring any human guidance on how to solve problems step by step. Specifically, comparing the performance of ChatGPT using \codename versus the best human-written CoT prompts from \citet{suzgun2022challenging}, \codename achieves consistently higher scores on all tasks.

\looseness=-1
Next, we compare \codename with self-consistency~(SC) decoding~\citep{wang2022self}. CoT+SC improves over CoT on two of the five tasks, but the improvements are not consistent. By contrast, \codename consistently outperforms CoT+SC by 2--26 points on all five tasks.

Additionally, we compare \codename with existing prompt optimization algorithms. 
APO improves over zero-shot prompting on three out of five tasks but underperforms it on the two tasks where the model needs to search through a wide range of strategies to find effective solutions. 
By contrast, \codename consistently outperforms zero-shot and CoT prompting, and improves over APO by~20--43 points on all five tasks.
When compared against Auto-CoT~\citep{zhang2022automatic}, \codename also archives higher accuracy by~+11 points on average. In summary, \codename outperforms strong decoding and prompt optimization baselines by~11--33 points on average.

\looseness=-1
Comparing the performance of \codename on different LLMs, we observe that InstructGPT underperforms ChatGPT on most tasks. However, we show that by using ChatGPT just as the initialization model~$LLM_1$ to bootstrap InstructGPT as $LLM_2$ in \codename, we can improve performance over InstructGPT alone by 5--71 points and achieve competitive or even better performance than ChatGPT alone on two of the five tasks. We show in the Appendix why that is: while InstructGPT can follow a given recipe and even be used for recombining and evolving them, it is less capable of generating diverse initial solutions in a zero-shot manner. However, through \codename, we can use ChatGPT to ``teach'' InstructGPT diverse strategies for solving the training problems, which are then recombined and evolved by InstructGPT into better CoT prompts for itself.

\looseness=-1
Furthermore, Table~\ref{tab:additional_results} shows the performance of \codename against zero-shot, few-shot and CoT prompting~(all using ChatGPT) on the remaining~15 tasks.\footnote{Based on the main results in Table~\ref{tab:main_results}, CoT+SC and Auto-CoT are more complicated than CoT but only slightly improves over CoT. Thus, we select CoT as a baseline here.}
\codename still outperforms zero-shot and few-shot prompting consistently and substantially by 14-15 points on average. Compared with CoT, \codename achieves better performance on 11 out of 15 tasks. On average, \codename outperforms CoT by +8.2 points. 
Interestingly, on tasks where CoT even underperforms zero-shot prompting, such as Movie Recommendation, Salient Translation Error Detection, and Word Sorting, \codename still improves over zero-shot prompting by large margins. This suggests that not all CoT recipes improve model performance, and some may even lead to degradation. This further emphasizes the need for algorithms like \codename for discovering and optimizing the CoT prompt to best exploit and compare LLMs.

Overall, these findings highlight the potential of \codename as a powerful method for automating CoT prompting on a wide range of tasks.

\begin{figure*}[ht]
    \begin{subfigure}[b]{0.33\textwidth}
        \includegraphics[width=\textwidth]{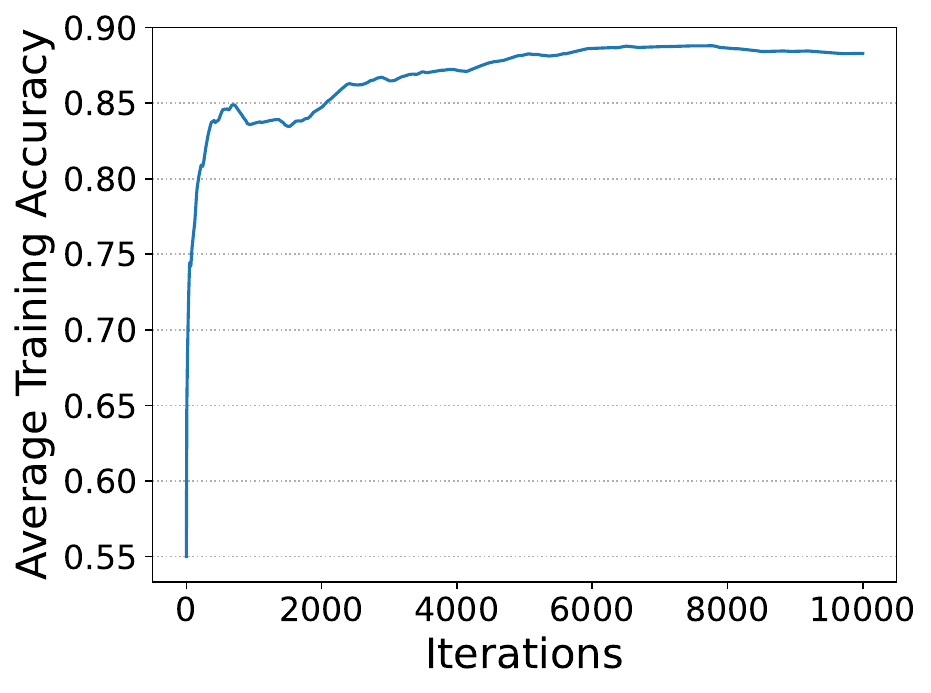}
        \caption{InstructGPT}
    \end{subfigure}
    \begin{subfigure}[b]{0.33\textwidth}
        \includegraphics[width=\textwidth]{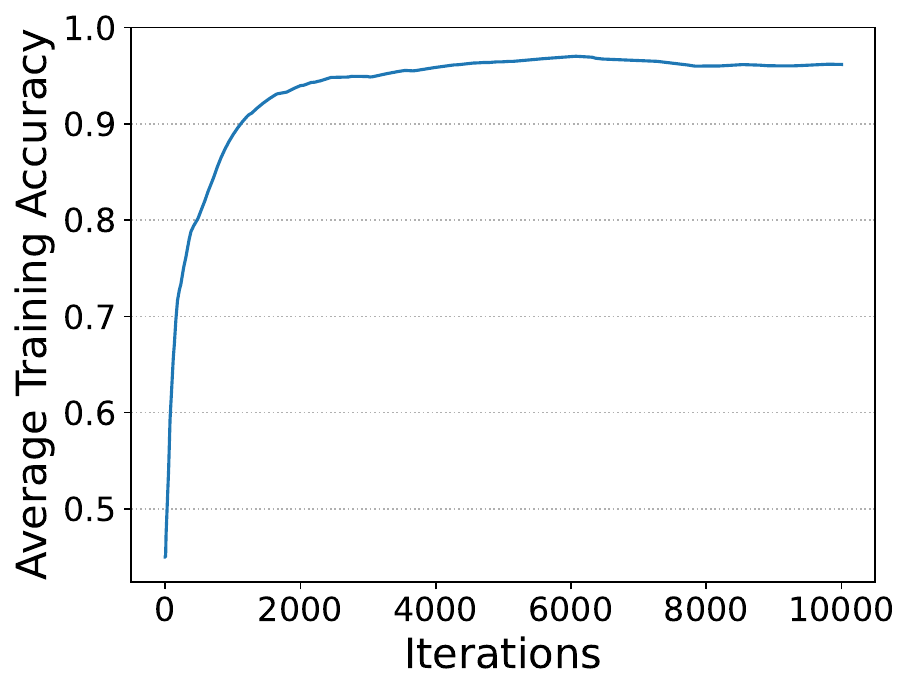}
        \caption{ChatGPT}
    \end{subfigure}
    \begin{subfigure}[b]{0.33\textwidth}
        \includegraphics[width=\textwidth]{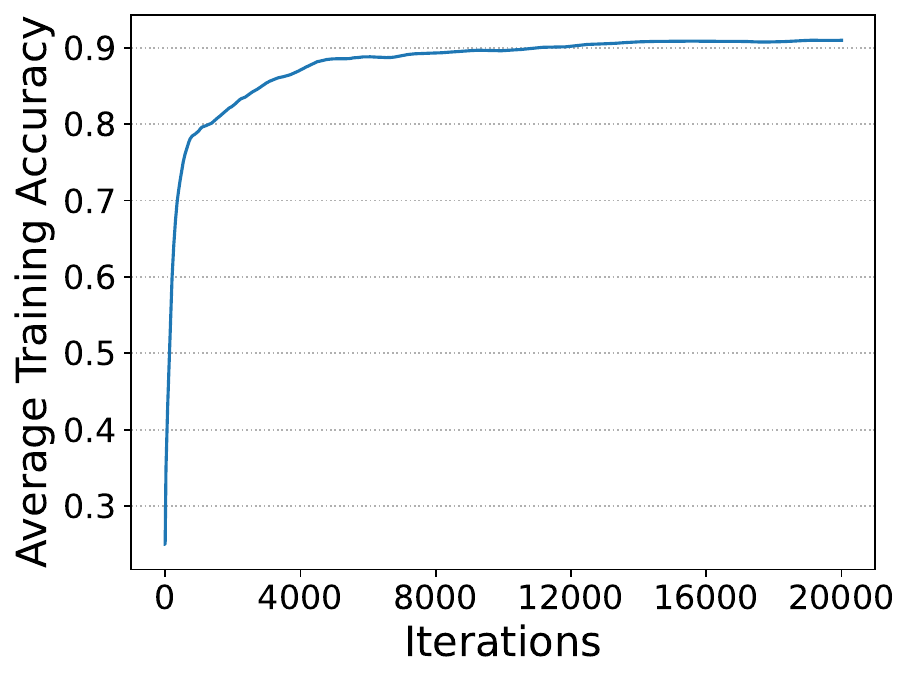}
        \caption{ChatGPT+InstructGPT}
    \end{subfigure}
\caption{Learning curves of the \codename algorithm using InstructGPT, ChatGPT, and the combo ChatGPT+InstructGPT models on the \textit{Logical Deduction} task. The y-axis shows the accuracy on training samples averaged over the current and all previous iterations.}
\label{fig:learning_curves}
\end{figure*}

\subsection{Quantitative Analysis}
\paragraph{Ablation Study}
\looseness=-1
We conduct an ablation study on the rejection sampling and recombination process. Results in Table~\ref{tab:ablation} show that, without rejection sampling, the test performance degrades substantially by 25 point on average. Always rejecting solutions that lead to incorrect answers also causes a degradation of 8 point. Additionally, not allowing multiple solutions to be recombined when sampling new solutions at the iterative sampling stage also hurts performance.

\begin{figure*}[ht]
    \centering
    \includegraphics[width=.9\textwidth]{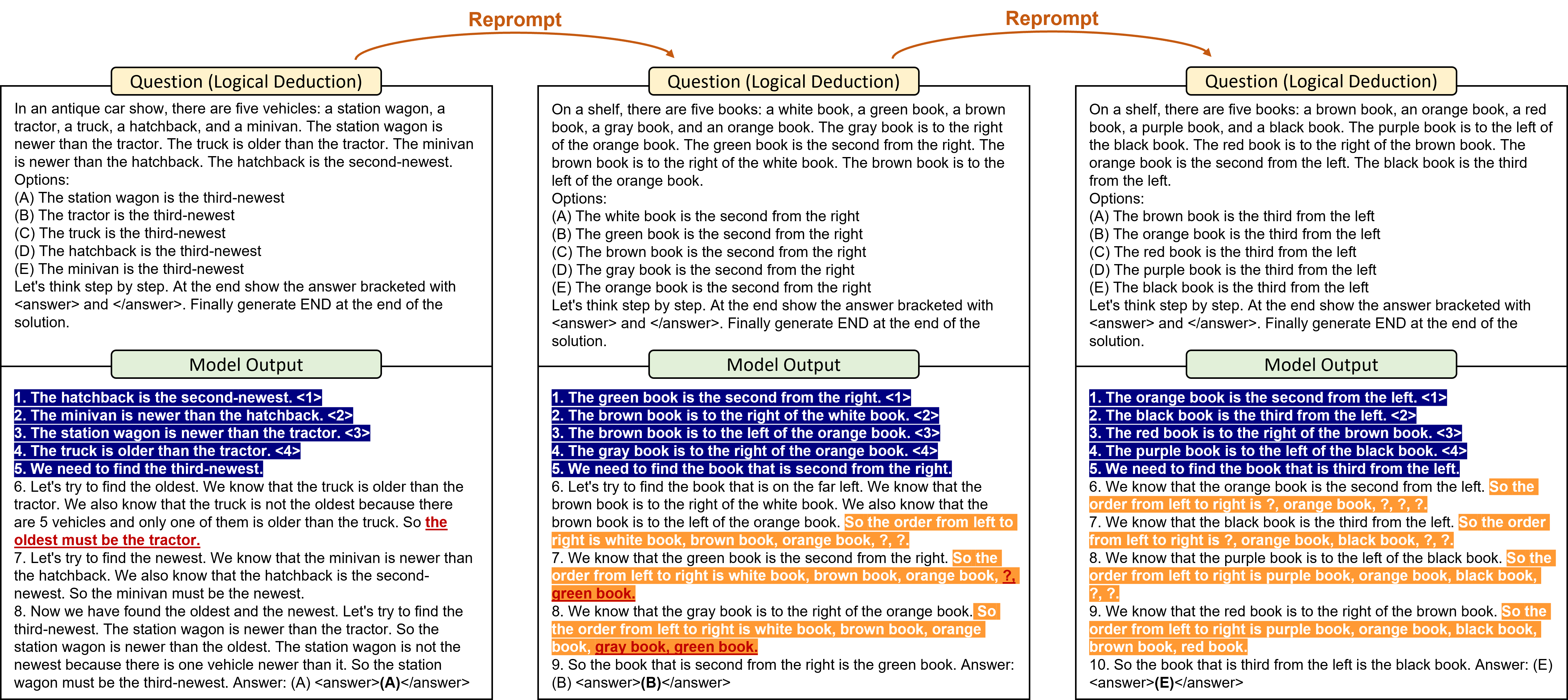}
\caption{An example of how the CoT recipes evolve through \codename. In the left-most recipe, the model~(ChatGPT) first reorders the constraints so that the ones with absolute ranking positions are considered prior to the ones with relative positions (highlighted in dark blue). Next, the model attempts to deduce the objects at specific positions but makes a mistake (see the red underlined part). Despite the error, this recipe still provides a useful strategy for solving similar problems \---\ when it is used in a prompt to solve another problem, the model first adopts the same strategy to reorder the constraints and then proposes another way to deal with the constraints (highlighted in orange). Although the resulting solution still contains errors, it makes a good recipe for solving this type of problem. Thus, when using it in a new prompt to solve yet another problem, the model can follow the same recipe and deduce the correct answer.}
\label{fig:reprompt_examples}
\end{figure*}

\begin{figure*}[ht]
    \centering
    \includegraphics[width=.48\textwidth]{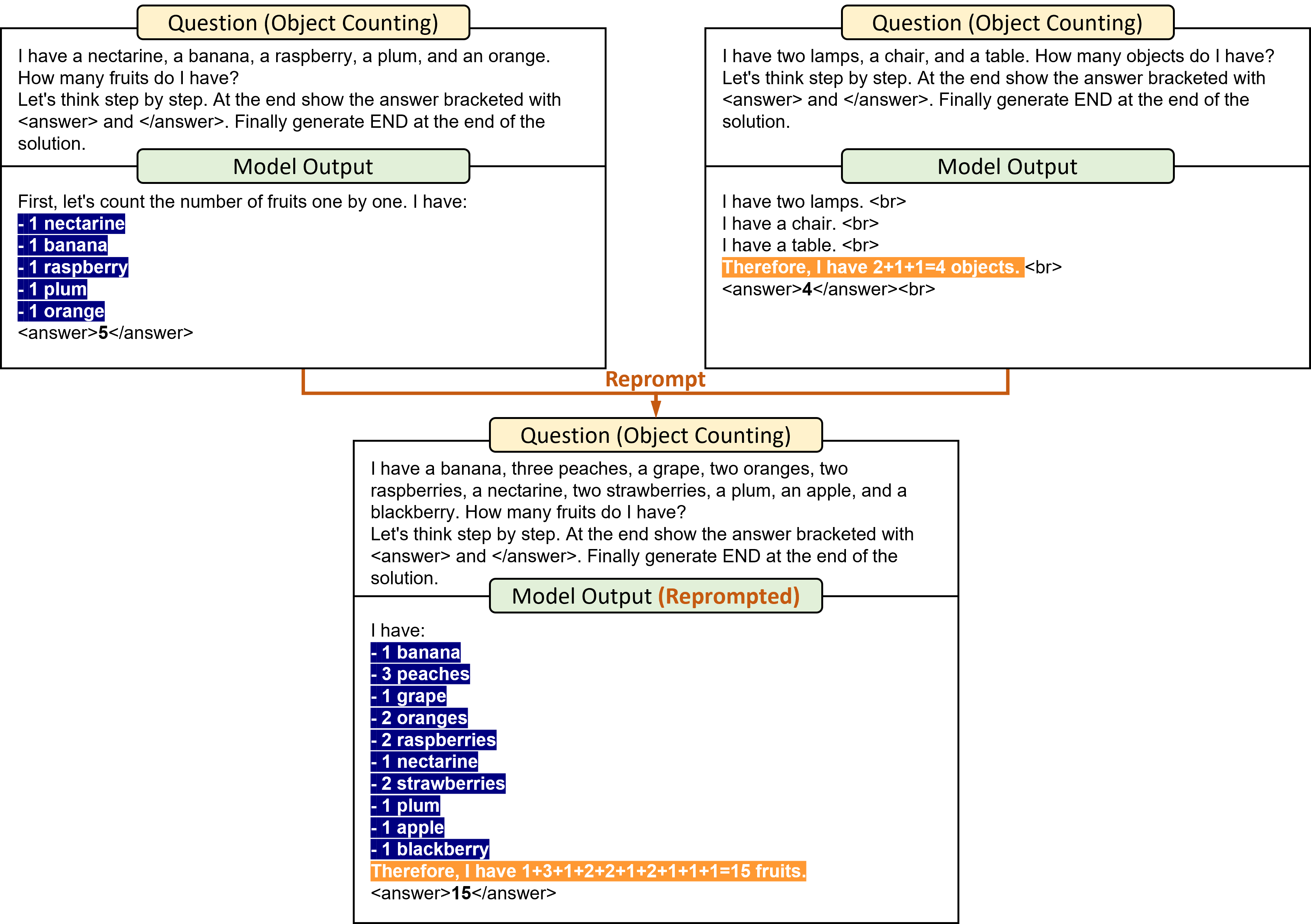}
    \includegraphics[width=.48\textwidth]{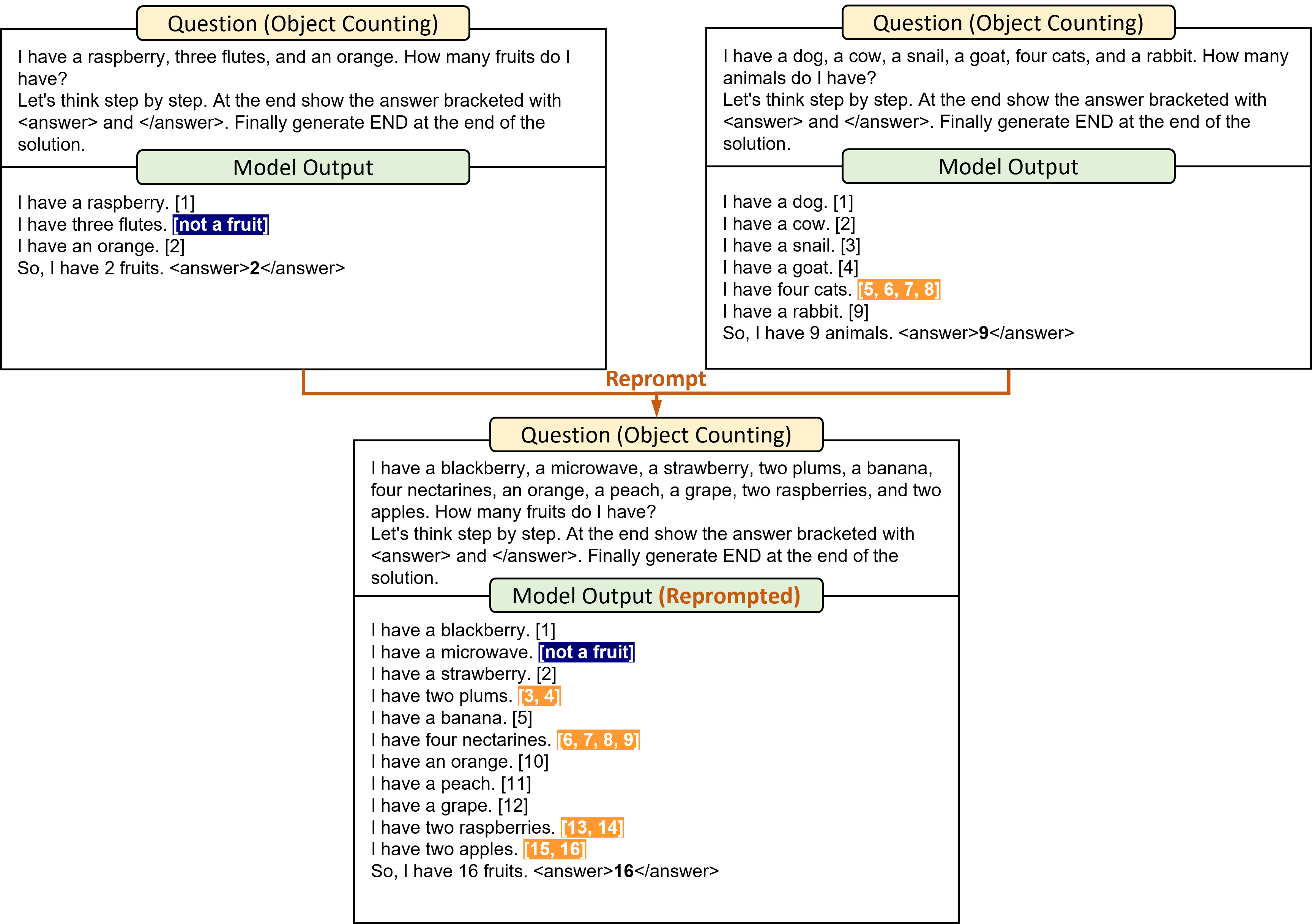}
\caption{Examples of how fragments from different recipes in a prompt can be (re)combined into a better recipe to solve a new problem through \codename.}
\label{fig:reprompt_combine_examples}
\end{figure*}

\paragraph{Do the generated CoT recipes generalize across models?} 
We test the best-performing CoT recipes optimized with InstructGPT, ChatGPT, or InstructGPT+ChatGPT through \codename on both InstructGPT and ChatGPT. 
As shown in Table~\ref{tab:test_across_models}, the CoT recipes optimized for one model may not work as well for other models.
Specifically, we observe that on tasks such as \textit{Logical Deduction} and \textit{Object Counting}, the best CoT recipes achieve similar performance on both InstructGPT and ChatGPT. However, on \textit{Geometric Shapes} and \textit{Temporal Sequences}, the best CoT prompts optimized for $LLM_2$ work well on $LLM_2$, but poorly with the other LLM \---\ using them on the other LLM leads to 18--19 points lower accuracy than testing with $LLM_2$~(see examples in Figure~\ref{fig:temporal_instructgpt_chatgpt}). On such tasks, using the prompt optimized for the testing LLM improves accuracy by 11--12 points over the same testing LLM with prompt optimized for other LLMs.
These results suggest that, to make a fair comparison between different LLMs, one needs to optimize the CoT prompt for each model.

\paragraph{\codename improves CoT recipes over iterations.} 
\looseness=-1
In Figure~\ref{fig:learning_curves}, we plot the average training accuracy~(averaged over iterations up to the current iteration) over training iterations on \textit{Logical Deduction}. For all three model variants, the initial training accuracy is relatively low, but it gradually increases (with occasional fluctuations) over iterations until convergence. This is the result of evolution and recombination of the recipes associated with training examples.

\paragraph{Compute and Resources}
We use the OpenAI APIs for all our experiments.\footnote{\url{https://platform.openai.com/docs/api-reference?lang=python}} Running \codename costs around~\$80 (in US dollars) on gpt-3.5-turbo and~\$800 on text-davinci-003 based on the standard pricing,\footnote{\url{https://openai.com/pricing}} while being exempted from any human cost.
By contrast, CoT prompting requires manual prompt construction and engineering, which costs not only human labor~(including the cost for humans to get familiar with the task itself and how LLM prompting works, write down various CoT solutions for each problem, test and optimize the solutions on the LLM) but also LLM queries, but these costs are typically neglected in previous works. In addition, previous works typically compare different LLMs using the same CoT prompt. While this strategy avoids additional costs for custimizing CoT prompt for each LLM~(even with \codename, one can also save the cost by running it with ChatGPT and using the inferred CoT prompt on other LLMs), it risks making unfair comparisons as we have shown in Table~\ref{tab:test_across_models} that the CoT prompt that works well on one model may be sub-optimal for another.

\subsection{Qualitative Analysis}
We observe that \textbf{even model outputs containing errors and unreasonable deductions can evolve into a high-quality recipe through \codename.} This is illustrated by the \textit{Logical Deduction} example in Figure~\ref{fig:reprompt_examples}, when $K=1$, where the model initially generates a recipe that is erroneous and contains illogical deductions. However, when this recipe is used as the new prompt for solving a similar problem, the model is able to exploit parts of the recipe and propose an alternative way to continue reasoning. Although the subsequent recipe still contains errors, it aids the model in correctly solving other problems when incorporated into a prompt. As a result, such recipes will be populated on other training samples, while the recipes that lead to low accuracy will eventually die out.

\paragraph{\codename combines fragments from different recipes into a better one.} 
\looseness=-1
\codename benefits from having multiple examples in the prompt, which allows the model to integrate various segments from different prompt recipes into a new recipe. As illustrated by the \textit{Object Counting} examples in Figure~\ref{fig:reprompt_combine_examples}, the model can combine large segments of reasoning steps, as well as small segments that address distinct cases to solve a more complex problem.
The resulting prompts sometimes, but not always, share similarities with the human-written prompts (See the Appendix).
\section{Related Work}
\paragraph{In-Context Learning} is an emergent ability of LLMs as they scale up in model sizes and training data, where an LLMs can learn to perform a task from a few examples in the context~(which is also referred to as few-shot prompting)~\citep{gpt3}. It has been shown to achieve promising few-shot and even zero-shot performance on various natural language processing~\citep{gpt3,schick2020exploiting,perez2021true} and program synthesis~\citep{austin2021program} tasks.

\paragraph{Reasoning via Chain-of-Thought Prompting} 
Chain-of-Thought~(CoT) prompting is a technique that enables LLMs to perform complex reasoning tasks by prompting them with a few examples with step-by-step solutions~\citep{wei2022chain, suzgun2022challenging}. CoT prompting has been shown to improve performance on various reasoning tasks, such as arithmetic reasoning~\citep{wei2022chain,zhou2022least}, symbolic reasoning~\citep{wei2022chain,zhou2022least}, multi-hop question answering~\citep{press2022measuring,arora2022ask}, and natural language inference~\citep{wang2022self}. However, designing effective CoT prompts requires human experts with an understanding of both the task and the prompting technique~\citep{zamfirescupereira2023why}, which limits the scalability and generalizability of CoT prompting.

Several works have attempted to \textbf{automate the process of CoT prompt discovery}. \citet{zhang2022automatic} proposed Auto-CoT, which uses LLMs to generate CoT solutions for diverse training questions in zero-shot and integrates the generated CoT solutions in the prompt for solving test questions. This method differs from \codename in that:~1) it focuses on the unsupervised setting and exploits a large set of example questions without annotated answers, and 2) it relies more heavily on the correctness of the zero-shot recipes as it does not have any iterative algorithm (as in \codename) to further improve the recipes. In our experiments, we adapted Auto-CoT to the few-shot setting and showed that \codename outperforms the few-shot version of Auto-CoT. 

\citet{deng2022rlprompt,zhang2023tempera} proposed to train an additional policy model to find the best prompt through reinforcement learning, but their approaches are limited to prompt optimization within a relatively small search space~(i.e. it is restricted to the prompts that are either extremely short or within a small edit distance from an initial prompt). \citet{zhou2023large} proposed a method for automatically generating, scoring and selecting effective instruction messages~$m$ for zero-shot chain-of-thought reasoning, which is orthogonal and can be potentially combined with our algorithm. \citet{paranjape2023art} introduced a framework that automatically retrieves demonstrations of related tasks from a task library and generates CoT solutions for the new task. However, this framework still requires collective human efforts to write demonstrations for a diverse set of tasks in the task library. In contrast, our \codename algorithm enables LLMs to solve complex reasoning tasks without any human guidance. Additionally, \citet{yoran2023answering} proposed a multi-chain reasoning~(MCR) method that prompts LLMs to combine pieces of information from multiple chains of thought to predict the final answer, which differs from our method in two ways: first, MCR combines multiple CoT solutions to the same question at test time, while \codename combines CoT solutions generated for different training questions before testing; second, MCR combines solutions only once, whereas \codename iteratively samples new solutions and recombines them. As a result, \codename generates effective CoT recipes from only a few training examples, resulting in improved test performance without slowing down test inference.

\section{Conclusion}
\looseness=-1
We introduce \codename, an automated prompt inference algorithm which, without human effort, discovers effective chain-of-thought~(CoT) prompts for each task given a few question-answer pairs. Experiments on~20 challenging reasoning tasks show that \codename achieves~+9.4 higher accuracy than human-written CoT on average. It also outperforms self-consistency decoding and the state-of-the-art prompt optimization algorithms by~11--33 points on average. 
Our results also suggest that LLM comparisons can be highly sensitive to the choice of CoT prompts, further emphasizing the need for automatic prompt discovery and optimization using algorithms such as \codename.

\section*{Acknowledgements}

We thank Bill Dolan, Sudha Rao and the reviewers for their valuable feedback.

\section*{Impact Statement}

This paper presents work whose goal is to advance the field of Machine Learning. There are many potential societal consequences of our work, none which we feel must be specifically highlighted here.

\bibliography{LLM,anthology}
\bibliographystyle{icml2024}

\newpage
\appendix
\onecolumn
\setcounter{figure}{0} 
\renewcommand{\thefigure}{\thesection.\arabic{figure}} 


\section{Additional Illustrations}

\paragraph{On sensitivity to initialization} 

\begin{figure}[t]
    \centering
    \includegraphics[width=0.7\textwidth]{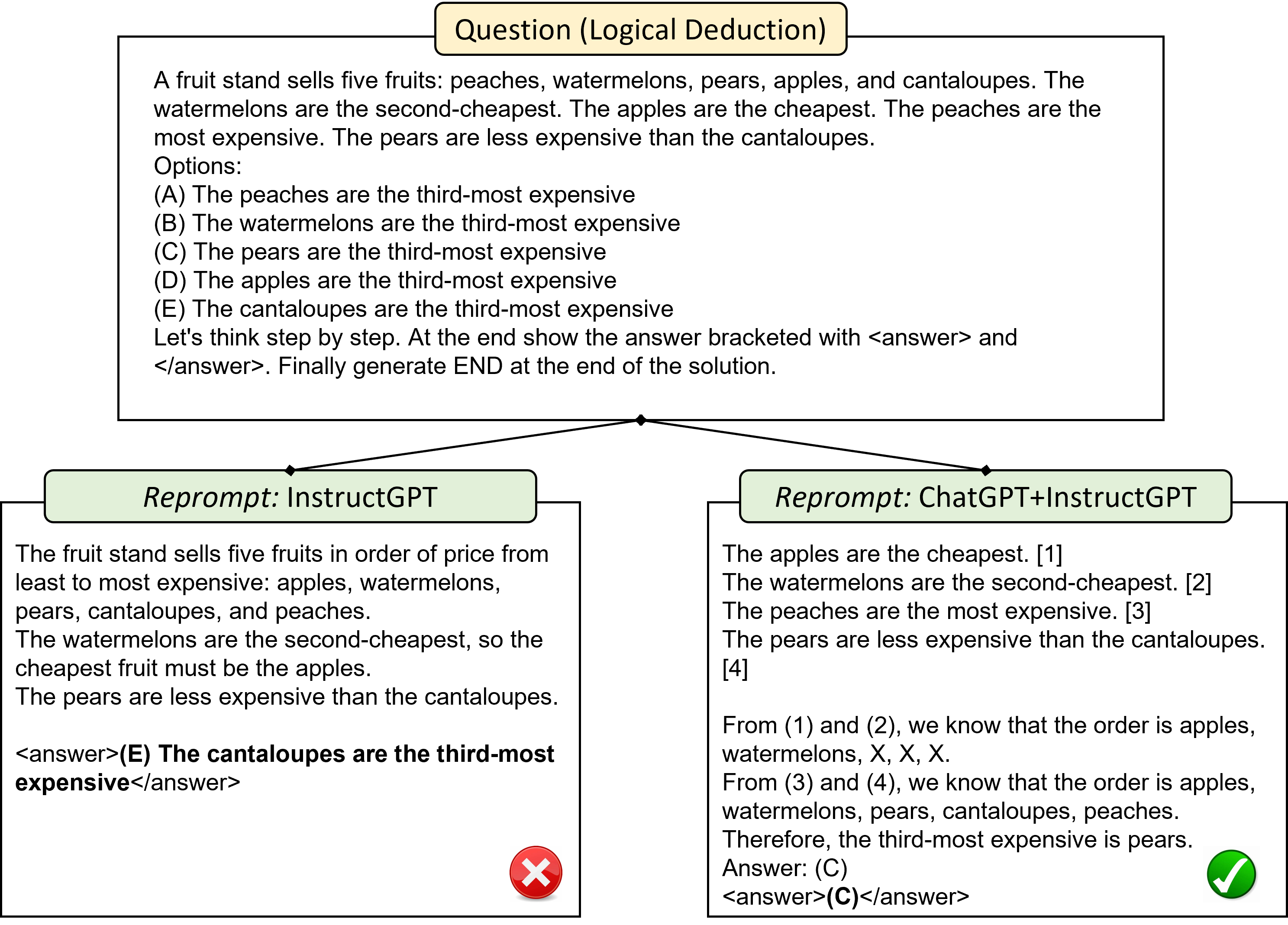}
\caption{Comparing the CoT recipes inferred through \codename using IntructGPT alone versus ChatGPT (for initialization) + InstructGPT (for sampling).}
\label{fig:reprompt_ChatGPT_InstructGPT}
\end{figure}

We have shown that \codename can be sensitive to initial recipe generation. Armed with the optimal prompts discovered with ChatGPT+InstructGPT through \codename, InstructGPT can reach test accuracy equalling or besting ChatGPT on most challenging reasoning tasks. However, on some tasks, such prompts could not be discovered using InstructGPT itself as the initialization model~$LLM_1$. Figure~\ref{fig:reprompt_ChatGPT_InstructGPT} points to a likely explanation: ChatGPT can generate a wider range of useful recipes, and whether these initial recipes lead to the correct solution or not, InstructGPT can follow them and, through \codename, refine and correct them iteratively. Thus, as we have  shown in our experiments, with a diverse pool of initial recipes, LLMs that may appear inferior based on their zero-shot performance may end up performing just as well or better than LLMs whose zero-shot performance is more encouraging. It would be interesting to see if \codename can use a mixture of LLMs in initialization to perform even better, or if humans can be put back into the loop to provide some initial recipes or some generic instructions on how to generate such recipes.

\paragraph{On transferability of discovered recipes} 

\begin{figure*}[t]
    \centering
    \includegraphics[width=0.7\textwidth]{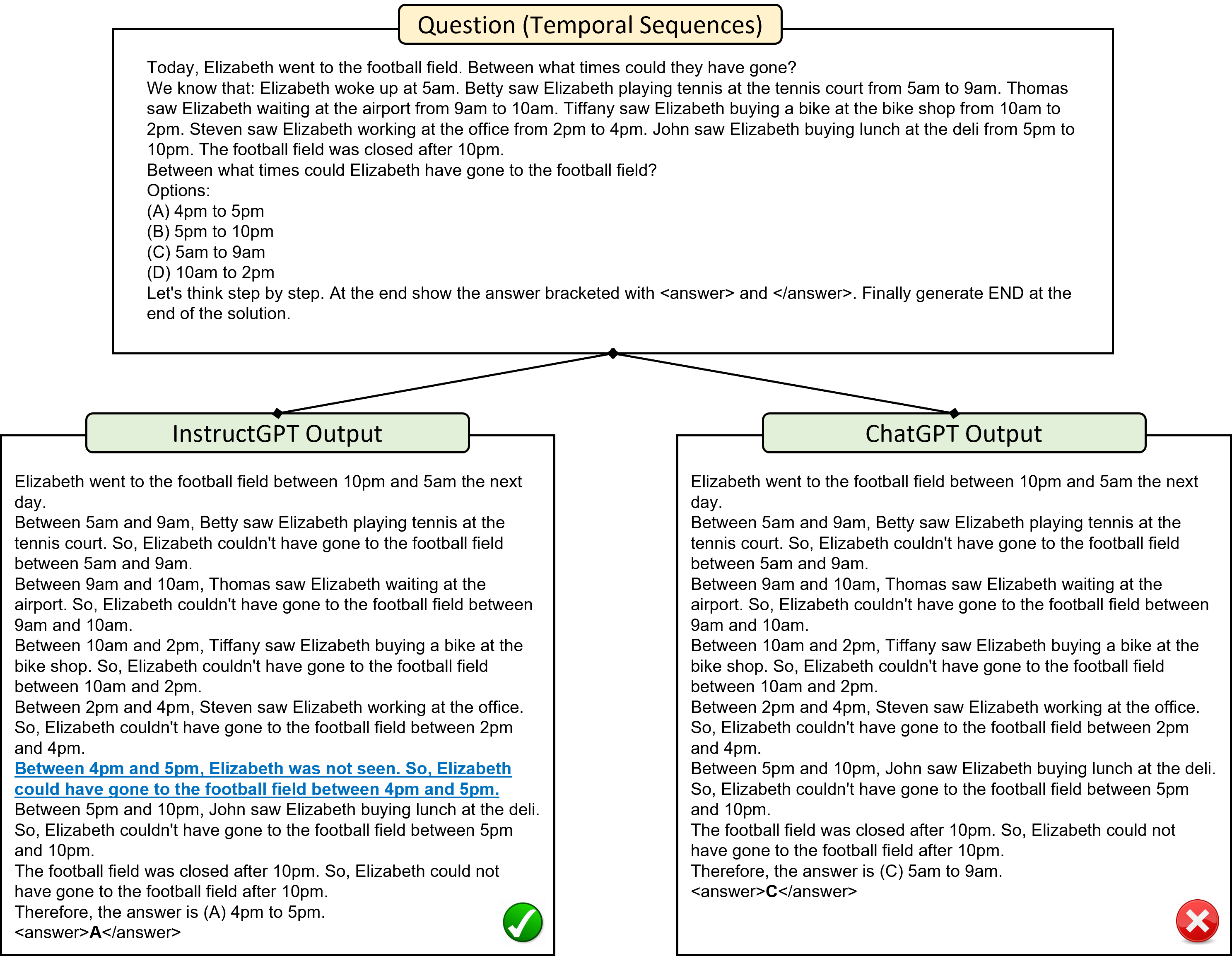}
\caption{An example on \textit{Temporal Sequences (BBH)} where ChatGPT underperforms InstructGPT using the same CoT prompt optimized for InstructGPT via \codename~(using ChatGPT+InstructGPT). ChatGPT fails to correctly execute the recipe as it skips a key step~(the blue underlined text from InstructGPT) to reach the final answer. (The illustration does not show the full prompt that precedes the puzzle~$x$ for brevity; it consists of~$5$ training examples with worked-out solutions that all follow the same strategy of solving these types of problems.)}
\label{fig:temporal_instructgpt_chatgpt}
\end{figure*}

The fact that $LLM_1$ (ChatGPT) can point $LLM_2$ (InstructGPT) in the right directions for prompt discovery does not mean that the discovered prompts, having been optimized for training performance on $LLM_2$, will perform well when used to prompt $LLM_1$. In fact, Table~\ref{tab:test_across_models} indicates that the discovered CoT recipes that work for one model may not necessarily work for other models. For example,  in the case of \textit{Temporal Sequences}, the best performance is achieved with a prompt trained with InstructGPT (after initialization with ChatGPT as $LLM_1$). But when using that prompt on ChatGPT, the test performance is by $18\%$ lower. Figure~\ref{fig:temporal_instructgpt_chatgpt} illustrates how ChatGPT and InstructGPT follow the same CoT prompt differently. Following the prompt recipes, the time intervals that need to be reasoned over are sorted, and among the sorted list, the missing interval was inserted as the possible interval when the person in question could have performed an activity. InstructGPT follows this procedure with accuracy over $99\%$, but ChatGPT sometimes skips the crucial line~(for this recipe) with the missing interval within the timeline and therefore obtains suboptimal test accuracy. However, the best performance of ChatGPT (using the CoT prompt optimized for itself through \codename) is only slightly lower than that of the ChatGPT+InstructGPT combination. 

These results suggest that, for a fair comparison between different LLMs, one needs to optimize the CoT prompt for each LLM using prompt optimization algorithms such as \codename.


\begin{figure*}[t]
    \centering
    \begin{subfigure}[b]{1\textwidth}
        \includegraphics[width=\textwidth]{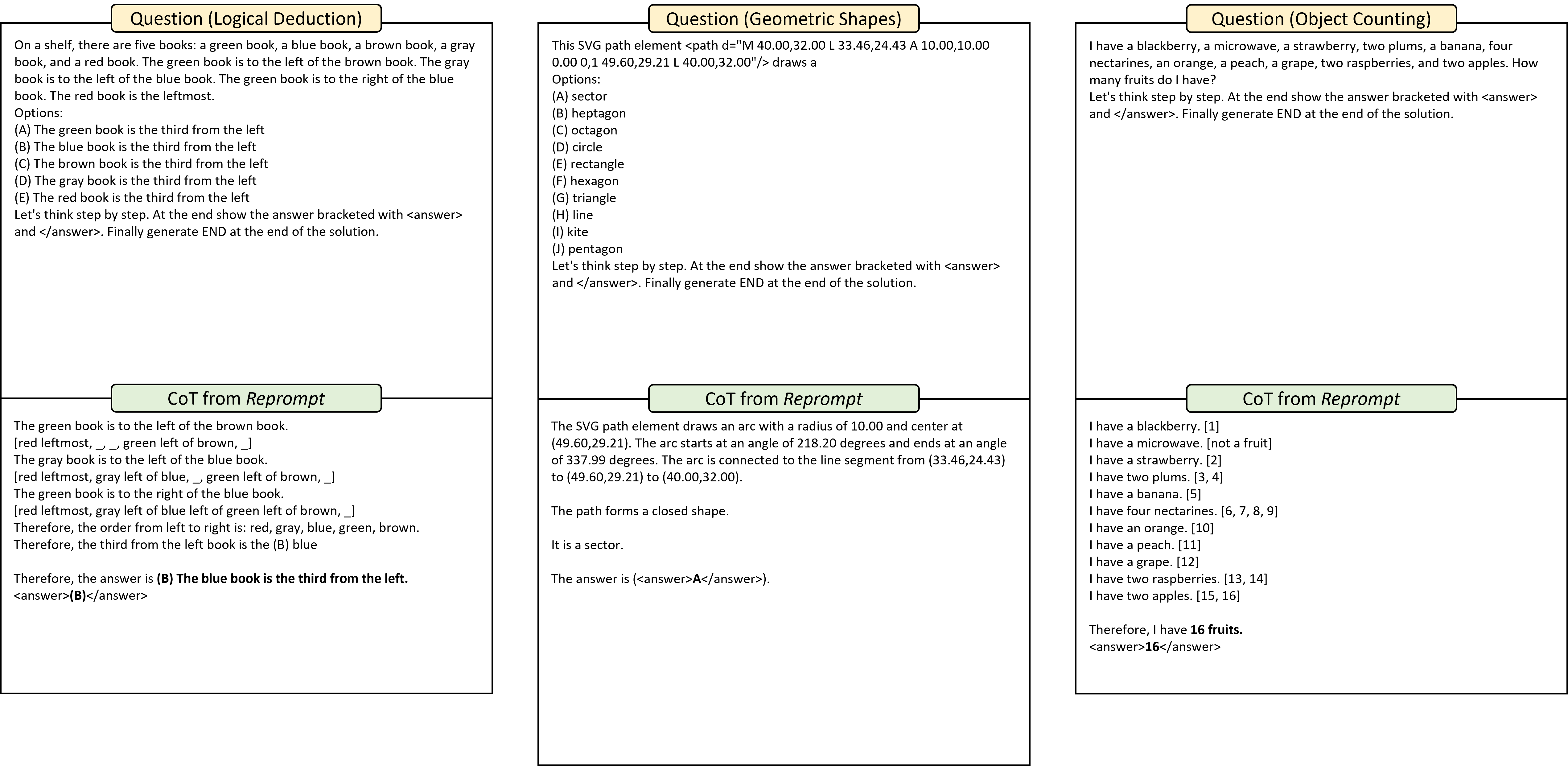}
        \caption{}
    \end{subfigure}
    \begin{subfigure}[b]{0.66\textwidth}
        \includegraphics[width=\textwidth]{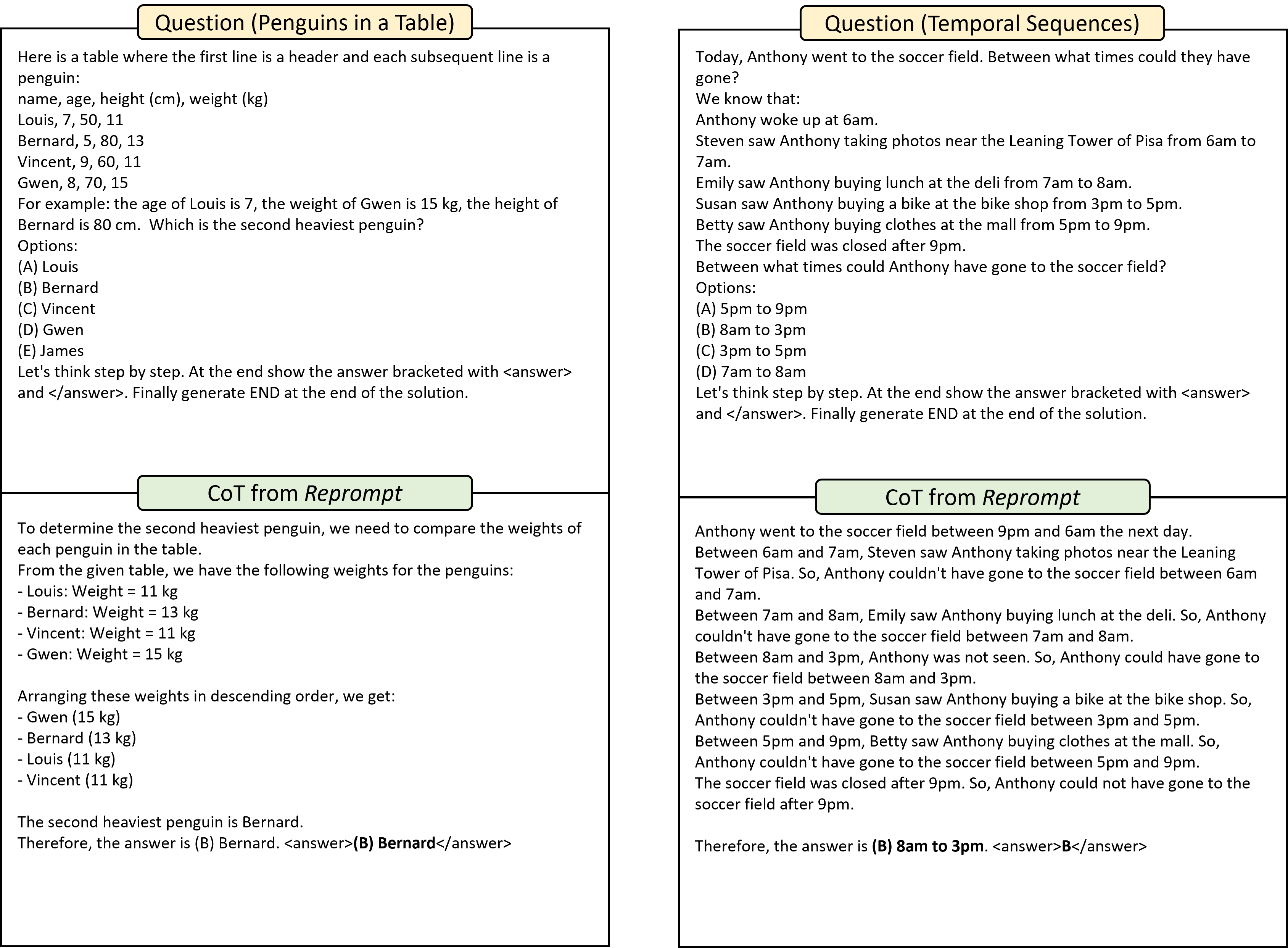}
        \caption{}
    \end{subfigure}
\caption{Examples of the best-performing CoT recipes inferred via \codename on \textit{Logical Deduction}, \textit{Geometric Shapes}, \textit{Object Counting}, \textit{Penguins in a Table}, and \textit{Temporal Sequences}.}
\label{fig:reprompt_final_CoT}
\end{figure*}
\paragraph{How do the model-generated CoT recipes differ from human-written ones?} 
\looseness=-1
In the paper, We evaluated the performance of the CoT prompt discovered through \codename and contrasted it with human-written ones. As illustrated by the example recipes in Figure~\ref{fig:reprompt_final_CoT}, the automatically discovered CoT recipes share some similarities to human-written ones on some tasks~(such as \textit{Logical Deduction}), but differs on other tasks. For instance, on \textit{Object Counting}, the CoT generated using \codename computes the total number of objects by incrementing the count one by one~(e.g. adding~$4$ to the count~$5$ by~``$[6, 7, 8, 9]$''), while in the human written recipe, it computes the addition through an arithmetic formula at the end. 

\end{document}